\documentclass[10pt,twocolumn,letterpaper]{article}

\usepackage{iccv}
\usepackage{times}
\usepackage{epsfig}
\usepackage{graphicx}
\usepackage{amsmath}
\usepackage{amssymb}
\usepackage{graphicx}
\usepackage{booktabs}
\usepackage{float}
\usepackage[section]{placeins}

\usepackage[breaklinks=true,bookmarks=false]{hyperref}

\iccvfinalcopy 

\begin{document}

\title{Domain Adaptive Monocular Depth Estimation With Semantic Information}

\author{
Fei Lu\\
The Robotics Institute\\
Carnegie Mellon  University\\
Pittsburgh, PA 15213, USA\\
{\tt\small flu2@andrew.cmu.edu}
\and
Hyeonwoo Yu\\
The Robotics Institute\\
Carnegie Mellon  University\\
Pittsburgh, PA 15213, USA\\
{\tt\small hyeonwoy@andrew.cmu.edu }

\and
Jean Oh\\
The Robotics Institute\\
Carnegie Mellon  University\\
Pittsburgh, PA 15213, USA\\
{\tt\small jeanoh@cmu.edu}
}

\maketitle

\begin{abstract}
   The advent of the deep learning has brought an impressive advance to monocular depth estimation, e.g., supervised monocular depth estimation has been thoroughly investigated. However, the large amount of the RGB-to-depth dataset may not be always available since collecting accurate depth ground truth according to the RGB image is a time-consuming and expensive task. Although the network can be trained on an alternative dataset to overcome the dataset scale problem, the trained model is hard to generalize to the target domain due to the domain discrepancy. Adversarial domain alignment has demonstrated its efficacy to mitigate the domain shift on simple image classification task in previous works. However, traditional approaches hardly handle the conditional alignment as they solely consider the feature map of the network. In this paper, we propose an adversarial training model that leverages semantic information to narrow the domain gap.
   Based on the experiments conducted on the datasets for the monocular depth estimation task including KITTI and Cityscapes, the proposed compact model achieves the state-of-the-art performance comparable to complex latest models and show favorable results on boundaries and objects at far distances.

\end{abstract}

\vspace{-10pt}
\section{Introduction}
\begin{figure}[t]
      \centering
      \includegraphics[scale=0.19]{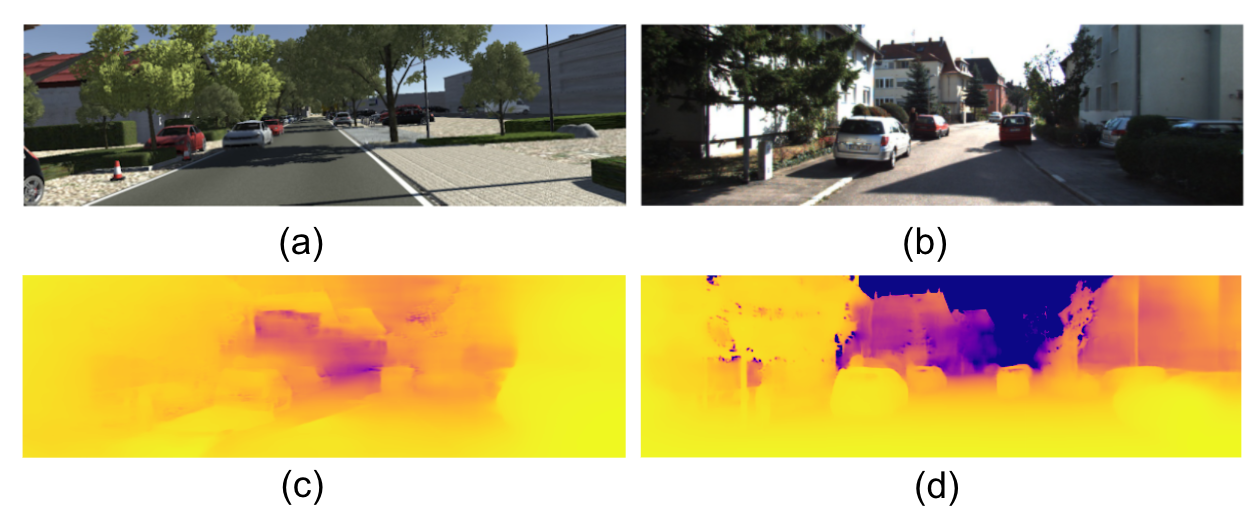}%
      \caption{An example of domain adaptation for monocular depth estimation task. (a) and (b) refer to source and target domain respectively. (c) is the predicted monocular depth before adaptation, (d) is after adaptation.}%
      \vspace{-10pt}%
      \label{figure1}%
\end{figure}
Depth map estimation from single RGB image is crucial for multiple tasks, e.g., autonomous driving, 3D reconstruction, virtual reality, augmented reality and down stream object detection task. Fully supervised monocular depth estimation methods such as Eigen Depth~\cite{eigen2014depth} and Make3D~\cite{Make3D} have shown promising results  with a large amount of precise RGB-to-depth dataset. However, collecting high quality depth ground truth requires high-precision calibrated sensors, e.g., expensive RGB-D camera and LiDAR sensors. RGB-D calibration is also a laborious task. Thus, deep learning-based monocular depth estimation approaches have shift the attention to self-supervised methods, e.g., Monodepth~\cite{monodepth},  Monodepth2~\cite{godard2019digging}, and EPC~\cite{EPC}. These self-supervised methods do not require high accuracy depth ground truth, instead, they only rely on synchronized stereo pairs or video sequences. 

All these aforementioned self-supervised methods have shown promising performance under standard benchmark, but image stereo pairs or video sequences may not be always available for a specific environment. To address this problem, alternative dataset with full annotations can be exploited for training the model, and the learned knowledge can be transferred to the desired environment by other domain alignment approaches. This domain adaptation technique can be apply to the monocular depth estimation \cite{AdaDepth} \cite{zheng2018t2net} \cite{lopezrodriguez2020desc}. Let $\mathcal{S}$ be the  fully-annotated source dataset and $\mathcal{T}$ be the non-annotated target dataset. Then our problem can be stated as bridging the domain gap between $\mathcal{S}$ and $\mathcal{T}$ for the monocular depth estimation. 

Meanwhile, the monocular depth estimation usually suffers from blurry edges, unnatural artifacts and loosing fine-grained details. Recent works \cite{geometrydepth,semanticallyguided} have shown that leveraging semantic information can improve monocular depth prediction by learning representations of scenes and objects. There exists an inverse relationship between object size and its depth for same class, a bike appears larger on 2D image plane when moving closer towards the camera. Also, there is an equivariance relationship for different classes, the bike appears smaller on 2D image plane compared to the car at the same depth. These approaches provide us a strong geometric cue to guide the depth estimation. Moreover, Recent studies found that utilize semantic information can also lead to narrow the domain discrepancy \cite{AdaDepth,kundu2019umadapt}. There exist a number of datasets with semantic annotations under various scenes, which shows a possibility to utilize the semantic resources for generalizing the method on different scenes. Therefore, we devise an semantic segmentation module by not only providing guidance to improve depth prediction, but also helping to narrow the gap from different domains.

\begin{figure}[t]
      \centering
      \includegraphics[scale=0.20]{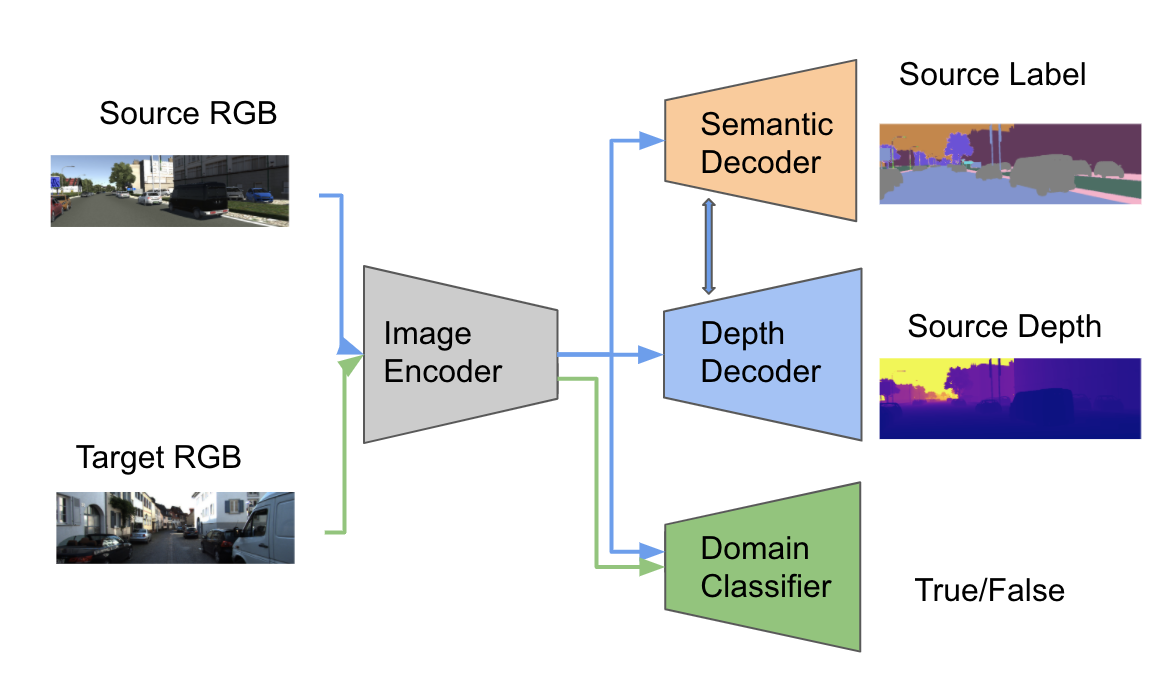}
      \caption{The proposed domain adaptation framework on monocular depth estimation task. Shared \textbf{Image Encoder} is used for domain feature extraction, \textbf{Depth Decoder} for source domain depth prediction and \textbf{Semantic Decoder} for geometric cue guidance.  Only ResNet Encoder and Depth Decoder are utilized at inference time. }%
      \label{figure2}%
      \vspace{-10pt}
\end{figure}

In this paper, we propose an end-to-end semi-supervised domain adaptation model \textit{Domain Adaptive Monodepth}, based upon state-of-the-art monocular estimation model Monodepth2\cite{godard2019digging}.  To address the conditional domain shift problem, we jointly train a depth estimation module, a semantic segmentation module and a domain adaptation module. We incorporate a domain adaptation module which contains a Gradient Reversal Layer (GRL) \cite{ganin2015unsupervised} and domain classifier together to minimize the domain discrepancy. We also employ the adversarial training manner to incentive depth estimation module and to exploit domain invariant representation simultaneously. To provide the geometric cue and refine the depth prediction results, we introduce semantic segmentation module which shares first few layers with depth prediction module. The depth and segmentation module are provided with fully annotated source data including depth and label ground truth; domain adaptation module only has non-annotated source and target data as inputs. The overview of the proposed model is illustrated in Figure  \ref{figure2}.

To summarize, the main contributions of this paper are the follows: 
\begin{itemize}
  \item we propose a semi-supervised adversarial monocular depth estimation model, which consists of ResNet encoder, depth decoder, semantic decoder and domain classifier. Domain classifier minimizes the H-divergence between source and target domains, enforces the ResNet encoder to learn domain invariant features. Semantic decoder motivates the depth decoder to learn more shape-based features instead of generic texture-based features;
  \item we introduce a multi-scale training loss for depth and segmentation decoder, calculating accumulated loss from small to big feature map, enriching the loss signal from coarse to fine-grained patch;
  \item we demonstrate effectiveness and robustness of our model on monocular depth estimation task by evaluating on widely used  KITTI\cite{kittiraw}, Virtual KITTI\cite{Gaidon:Virtual:CVPR2016} and Cityscapes\cite{cordts2016cityscapes} datasets.
\end{itemize}

\section{\textbf{Related Works}}

\subsection{\textbf{Monocular Depth Estimation}}

Monocular Depth Estimation is an ill-posed problem that suffers from scale ambiguity and scale shifting issues, e.g., an image of the same color pixels can be projected to the multiple depth values. To compensate the lack of multi-view information, global semantic information and local object information are generally utilized to predict accurate depths. Fully Convolutional Network (FCN)~\cite{eigen2014depth,laina2016deeper} is proposed to extract meaningful objectiveness information for fast and efficient depth estimation. There are two main streams in FCN-based methods to address this ambiguity problem: (1) supervised method that take pair-wise data of color image and ground-truth depth during training, which unquestionably demonstrates excellent performance; (2) unsupervised method that requires stereo color image pairs or consecutive temporal frames in the training time, but only needs single-view color image as an input at the inference or test time.

\subsubsection{\textbf{Supervised Monocular Depth Estimation}}
Eigen et al.~\cite{eigen2014depth} were the first to utilize CNNs for the depth estimation task by exploiting coarse  and fine-grained features with a two-scale architecture. During training, scale invariant error is used as loss function in order to fix scale drifting issue in monocular depth estimation. Bo et al.~\cite{7298715} further improve the depth prediction results by utilizing a hierarchical graphical model such as Conditional Random Fields (CRF) on deep features. Laina et al.~\cite{laina2016deeper} is the first group to present an encoder-decoder architecture that encompasses residual learning and discard the post-processing techniques used in CRF-based models.

\subsubsection{\textbf{Unsupervised Monocular Depth Estimation}}
Cl{\'{e}}ment Godard et al.~\cite{monodepth} present a training objective that exploits epipolar geometry constraints to warp a right image to a corresponding left image with the left-right consistency and the image reconstruction loss during training, resulting in plausible quality depth images. Later, \cite{godard2019digging} proposes another approach that treats temporal images in a sequence as multi-view observations, feeding the images before and after the keyframe into a pose network , demonstrating its effectiveness on the KITTI dataset.  


\subsubsection{\textbf{Depth and Semantic Relationships}}
Jiao et al.~\cite{Jiao_2018_ECCV} create a lateral sharing unit for the depth and semantic decoder, alongside the attention-driven loss for the depth and semantic outputs. Mousavian et al.~\cite{mousavian2016joint} propose a framework that jointly trained the network for depth and semantic prediction tasks with a shared backbone, to feed to a fully connected CRF to capture the contextual relationships and cues between the depth and semantic. Casser et al.~\cite{casser2018depth} impose an object size constraint for dynamic objects in the video sequences and refine the depth prediction in an online fashion.

\subsection{\textbf{Domain Adaptation}}
In general, top-performing deep architectures are trained on a great amount of annotated data. If the amount of annotated data is insufficient in a target domain, the traditional approaches usually take the domain adaptation techniques that use annotated data from a different domain but inherently shares the same underlying features, e.g., photo-realistic synthetic images. In the visual domain adaptation area, there are a few approaches: transferring the style from the source  to the target ~\cite{hoffman2018cycada}, enforcing the cycle consistency; employing adversarial training with pseudo labels to align either the intermediate features~\cite{ganin2015unsupervised,tzeng2017adversarial} or the final outputs of the domains~\cite{tsai2020learning}; or applying transformation or whitening techniques on the output predictions, regularizing with custom loss terms~\cite{Roy_2019_CVPR,sajjadi2016regularization}.

Various studies have explored the domain adaptation approaches for depth prediction task. Abarghouei et ~\cite{atapour2018real} propose a two-stage approach: the depth estimation module is firstly trained on synthetic data, then using style transfer to adapt to real data with cycle-consistency loss. Saito et al.~\cite{zheng2018t2net} also consider a two-stage network, known as $T^2 Net$, but they choose a unidirectional forward loss instead of a cycle-consistency loss for the image  translation network. Lopez Rodriguez et al.~\cite{lopezrodriguez2020desc} present the $DESC$ model that builds on $T^2 Net$, incorporating semantic, edge, and instance mask as the training network inputs. Jogendra et al.~\cite{AdaDepth} propose $AdaDepth$, an encoder-decoder framework with two shared ResNet-50 image encoders for source and target input images, two discriminators for latent features and final depth outputs. Our proposed model draws an inspiration from DANN~\cite{ganin2015unsupervised} that focuses on aligning latent features outputting from the encoder, and also leverages semantic information in the training time to guide to regress pixel-wise depths.


\section{\textbf{Approach}}

\begin{figure*}[ht]
      \center
      \includegraphics[width=0.8\linewidth]{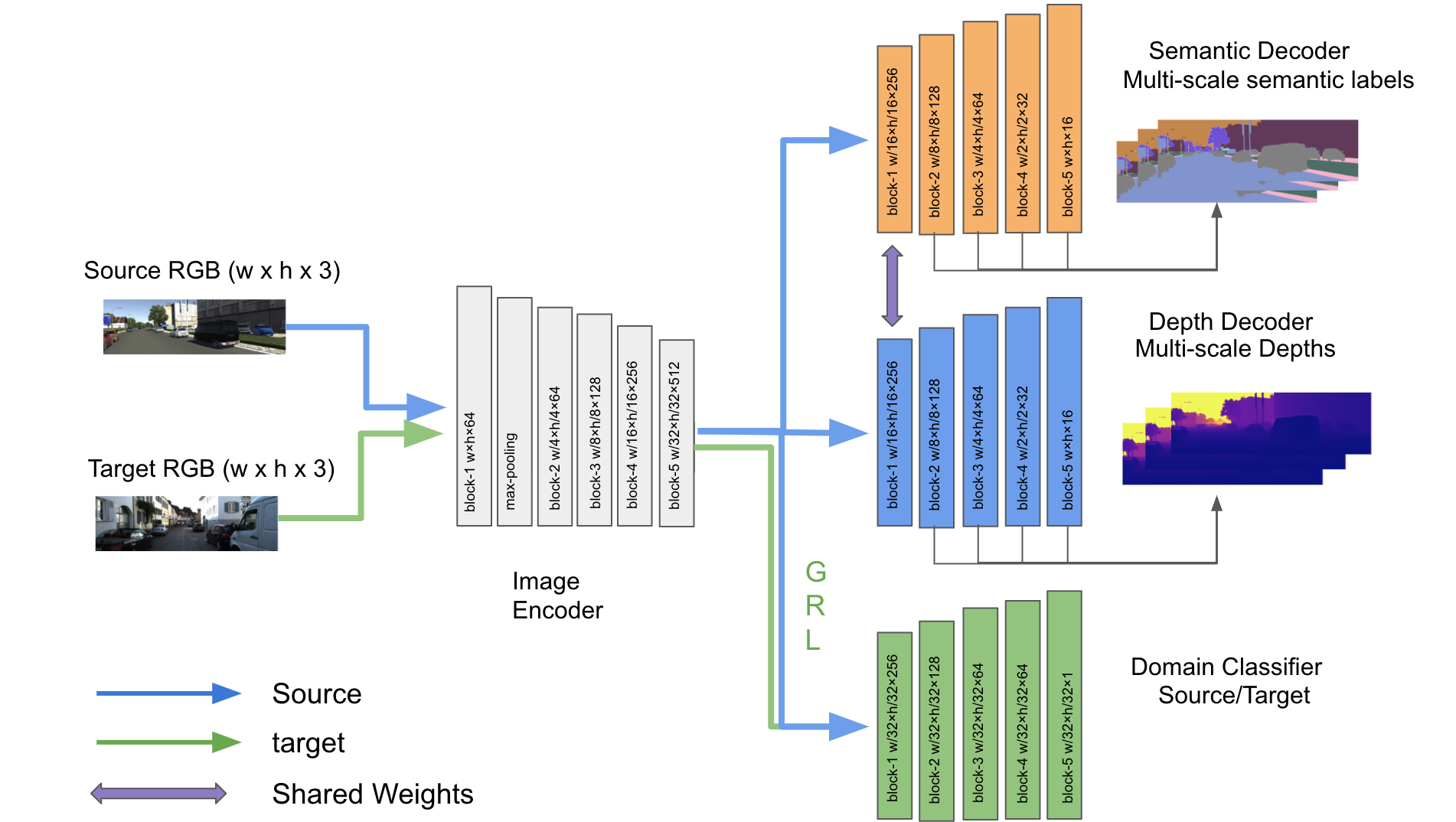}
      \caption{An overview of the proposed domain adaptation network for monocular depth estimation. \textbf{Image Encoder:} extracts source and target input image features; \textbf{Depth Decoder: } regresses pixel-wise depth map for source domain depth estimation; \textbf{Semantic Decoder: } predicts  semantic segmentation map meanwhile provides geometric cue for depth estimation; \textbf{Domain Classifier: } tends to remove domain specific information and align source and target latent feature space. The skip connections between encoder and decoders have been omitted in this graph. Size of output feature map is denoted as $ \textbf{width} \times \textbf{height} \times \textbf{channel}$ inside each block.}%
      \label{figure3}%
      \vspace{-11pt}
\end{figure*}

We present our adversarial monocular depth estimation model for domain adaptation with semantic information, an overview of the model structure is shown in Figure~\ref{figure3}. The semantic segmentation decoder and the domain classifier are auxiliary modules only used in the training time; we only utilize the image encoder and the depth decoder during the inference time on the target domain. Note that we omit the skip connections  for simplicity.

\subsection{\textbf{Latent Representation with a Domain Classifier}}\label{sec:domain-classifier}
Assuming that both the source images, denoted by $x_s \in X_s$, and the corresponding ground truths, $y_s \in Y_s$, are sampled from the source dataset $\mathcal{S}$ under the source distribution $p_{s}(x,y)$. Similarly, target images $x_t \in X_t$ are sampled from the target dataset $\mathcal{T}$ under the target distribution $p_{t}(x,y)$, where $p_{s} \neq p_{t}$. In our problem setting, the target ground truths $y_t \in Y_t$ are not provided at the training phase. 

Consider a typical encoder-decoder model for depth estimation, where the encoder $Enc(x;\theta_{enc})$ extracts input image features as latent representations, and decoder $Dec(x;\theta_{depth})$ exploit and expands the latent representations to depth. By defining the bottleneck latent representation as $z = Enc(x;\theta_{enc})$, we have predicted depth as the output of a decoder from the latent representation as: $ \hat{y} = Dec(z;\theta_{depth}) = Dec(Enc(x;\theta_{enc});\theta_{depth})$. If the encoder $Enc(x;\theta_{enc})$ and decoder $Dec(x;\theta_{depth})$ only have access to the source annotations (ground truths) during the training time, it can cause the model hard to generalize on a novel target dataset during the inference time. The shifting domain will result in different latent representations,  deteriorating depth prediction performance. Ganin et al.~\cite{ganin2015unsupervised} claim that using a Gradient Reversal Layer (GRL) and a domain classifier can make the latent representation domain-invariant and have shown the effectiveness on a simple classification task. Motivated by this idea, we attach a GRL and a domain classifier $ D(z; \theta_{domain}) $ to align the latent feature representations, so that the decoder can achieve a consistent performance under the same representation domain.
\subsection{\textbf{Depth Prediction with Semantic Information}}

A semantic segmentation map provides abundant shape features and structural information of a scene; such explicit information about the overall spatial structures can help to regress pixel-wise depths. Previous works~\cite{geometrydepth,semanticallyguided} have shown the effectiveness of semantic information in the supervised depth estimation settings. In addition, depth prediction also benefits the semantic label prediction task during training, yielding faster and accurate training results~\cite{chen2019learning,liu2019synthesis}. This synergistic relationship inspires us to incorporate a segmentation decoder along side the depth decoder during training.  

Intuitively, the lower level representation captures image-specific features, e.g., the intensity of colors, angle of edges, etc., whereas the higher level representation captures more general, abstract features, such as instance categories ``bed”, ``door”, etc. Specifically, in our depth estimation task, we want to learn the objectiveness information from the high level latent representations. Since semantic labels and depth maps are generated from the same latent representation with a joint distribution, we can borrow the weight-sharing idea as in CoGAN ~\cite{liu2016coupled}, enforcing weight sharing between the semantic decoder $Dec(x;\theta_{seg})$ and the depth decoder $Dec(x;\theta_{depth})$ on the first few layers, to speed up training and reduce the computation cost. 

\subsection{\textbf{Loss Function}}
\subsubsection{\textbf{Adversarial Loss}}

Our model outputs two latent representations, 
$ z_s = Enc\left(x_{s};\theta_{enc}\right)$ and 
$ z_t = Enc\left(x_{t};\theta_{enc}\right)$, with respect to the source and the target domains.
We define an adversarial loss, $\mathcal{L}_{adv}$, for the latent space representation as follow:
\begin{equation}
    \begin{aligned} 
     \mathcal{L}_{adv}=\mathbb{E}_{x_{s} \sim X_{s}}\left[\log D\left(Enc\left(x_{s};\theta_{enc}\right);\theta_{domain}\right)\right] \\
     +\mathbb{E}_{x_{t} \sim X_{t}}\left[\log \left(1-\left(D\left(Enc\left(x_{t};\theta_{enc}\right);\theta_{domain}\right)\right)\right)\right] 
    \end{aligned}
\end{equation}
where $\theta_{domain}$ is the parameters for the domain classifier described in Section~\ref{sec:domain-classifier}.
\subsubsection{\textbf{Multi-scale Depth Prediction Loss}}
We obtain 4-scale outputs (sizes of $d/8, d/4 , d/2, d$ with respect to input size $d$) from last 4 upsample-convolution depth decoder blocks. The total depth loss is the combination of loss from each scale. We borrowed the depth loss from Monodepth2 with different weights ~\cite{godard2019digging}, resulting in a conglomeration of SSIM loss $\mathcal{L}_{SSIM}$ ~\cite{russakovsky2015imagenet,loss_function}, L1 loss $\mathcal{L}_{l1}$, and edge-aware smoothness loss $\mathcal{L}_{smooth}$. 
Let $x_s$ and $y_s$ denote the input source image and the ground-truth source depth, respectively. The predicted depth is the output of the decoder defined as: $\hat{y_{s}} = Dec(z;\theta_{depth}) = Dec(Enc(x_s;\theta_{enc});\theta_{depth})$. The L1 loss is defined as absolute pixel-wise difference between predicted and ground truth depth:
\begin{equation}
\mathcal{L}_{l1} = \left\|\hat{y_{s}} - y_s\right\|_{1}
\end{equation}

Since the depth discontinuities usually happen on the edge of an object, we introduce a penalty term, $\mathcal{L}_{smooth}$, to optimize this problem as in ~\cite{godard2019digging}, where $y_{s}^{*}$ is a mean-normalized predicted depth:
\begin{equation}
\mathcal{L}_{smooth} = \left|\partial_{x} y_{s}^{*}\right| e^{-\left|\partial_{x} x_{s}\right|}+\left|\partial_{y} y_{s}^{*}\right| e^{-\left|\partial_{y} x_{s}\right|}
\end{equation}

Therefore, the total depth loss is defined as:
\begin{equation}
\mathcal{L}_{depth}  = (1 - \lambda_1)\mathcal{L}_{SSIM} + \lambda_1 \mathcal{L}_{l1} + \sum_{scale} \lambda_2 \mathcal{L}_{smooth}
\end{equation}
\subsubsection{\textbf{Multi-scale Segmentation Loss}}

Similarly, the semantic decoder also predicts multi-scale segmentation maps. Let $x_s$ denote the input source image; $y_s$, source segmentation label ground truth; and $\hat{y_{s}}$, predicted segmentation label defined as: $\hat{y_{s}} = Dec(z;\theta_{seg}) = Dec(Enc(x_s;\theta_{enc});\theta_{seg})$. We define our segmentation loss as a multinomial logistic loss, i.e., cross entropy loss for K classes:
\begin{equation}
\begin{aligned} \mathcal{L}_{seg} &=\mathbb{E}\left[-\sum_{i=1}^{scales}\sum_{j=1}^{\left|y_{s}\right|} \sum_{k=1}^{K} 1^{y^{j}_s=k} y_s^j \log \left(\hat{y_s^j}\right)\right] \end{aligned}
\end{equation}
where $\sum_{i=1}^{scales}$ means accumulating over all $i$ scales; $\sum_{j=1}^{\left|y_{s}\right|}$ represents the summation over all the pixels in the segmentation map; and $\mathbf{1}^{y^{j}_{s}=k}$ is a one-hot indicator at the j-th pixel from source image.
\subsubsection{\textbf{Overall Loss}}

The overall loss during training can be constructed as follows:
\vspace{-5pt}
\begin{equation}
    \mathcal{L} = \lambda_{adv}\mathcal{L}_{adv}  + \lambda_{depth} \mathcal{L}_{depth} + \lambda_{seg} \mathcal{L}_{seg}
\end{equation}

\section{\textbf{Experiments}}

\begin{table*}[t]
\begin{center}
\caption{Comparative results for \textbf{Virtual KITTI} to \textbf{KITTI} with median scaling. AdaDepth results are taken from official paper~\cite{AdaDepth}, T\textsuperscript{2}Net ~\cite{zheng2018t2net} and DESC~\cite{lopezrodriguez2020desc} results are computed with official pretrained weights. All the model are evaluated under the same evaluation code from Monodepth2~\cite{godard2019digging}.}
\label{table1}
\begin{tabular}{llllllll}
\toprule[0.5mm]
             & \multicolumn{4}{c}{Lower  $\downarrow$} & \multicolumn{3}{c}{Higher  $\uparrow$} \\
             \cmidrule(r){2-5}                        \cmidrule(l){6-8} 
             
Method      & Abs Rel &  Sq Rel &  RMSE &  RMSE log & $\delta < 1.25$  & $\delta < 1.25^{2}$ & $\delta < 1.25^{3}$ \\  
\midrule[0.5mm]

\textbf{cap = 50m} \\
AdaDepth~\cite{AdaDepth}                     & 0.203 & 1.734 & 6.251 & 0.284 & 0.687 & 0.899 & 0.958 \\
T\textsuperscript{2}Net ~\cite{zheng2018t2net}     & 0.165 & 1.034 & 4.501 & 0.235 & 0.772 & 0.927 & 0.972 \\
DESC (Img Trans. + Seg.) ~\cite{lopezrodriguez2020desc}           & 0.153 & 0.995 & 4.233 & 0.229 & 0.795 & 0.929 & 0.966 \\
DESC (Img Trans. + Seg. + Edge.) ~\cite{lopezrodriguez2020desc}   & 0.149 & \textbf{0.819} & \textbf{4.172} & 0.221 & 0.805 & 0.934 & 0.975 \\
DA-Monodepth (ours)           & \textbf{0.149} & 0.958 & 4.321 & \textbf{0.210} & \textbf{0.811} & \textbf{0.947} & \textbf{0.982} \\

\midrule[0.5mm]

\textbf{cap = 80m} \\
AdaDepth ~\cite{AdaDepth}                   & 0.214 & 1.932 & 7.157 & 0.295 & 0.665 & 0.882 & 0.950 \\
T\textsuperscript{2}Net ~\cite{zheng2018t2net}      & 0.173 & 1.396 & 6.041 & 0.251 & 0.757 & 0.916 & 0.966 \\
DESC (Img Trans. + Seg.)~\cite{lopezrodriguez2020desc}           & 0.171 & 1.332 & 5.818 & 0.250 & 0.771 & 0.918 & 0.966 \\
DESC (Img Trans. + Seg. + Edge.) ~\cite{lopezrodriguez2020desc}   & 0.156 & \textbf{1.067} & 5.628 & 0.237 & \textbf{0.787} & 0.924 & 0.970 \\
DA-Monodepth (ours)           & \textbf{0.156} & 1.272 & \textbf{5.118} & \textbf{0.221} & 0.783 & \textbf{0.927} & \textbf{0.972} \\

\bottomrule[0.5mm]
\end{tabular}
\end{center}
\end{table*}

\begin{table*}[t]
\begin{center}
\caption{Camparative results for \textbf{Cityscapes} to \textbf{KITTI} with median scaling. T\textsuperscript{2}Net ~\cite{zheng2018t2net} and DESC~\cite{lopezrodriguez2020desc} results are computed with official pretrained weights.}
\label{table2}
\begin{tabular}{llllllll}
\toprule[0.5mm]
             & \multicolumn{4}{c}{Lower  $\downarrow$} & \multicolumn{3}{c}{Higher  $\uparrow$} \\
             \cmidrule(r){2-5}                        \cmidrule(l){6-8} 
             
Method      & Abs Rel &  Sq Rel &  RMSE &  RMSE log & $\delta < 1.25$  & $\delta < 1.25^{2}$ & $\delta < 1.25^{3}$ \\
\midrule[0.5mm]
\textbf{cap = 80m} \\
T\textsuperscript{2}Net ~\cite{zheng2018t2net}          & 0.173 & 1.335 & 5.640 & 0.242 & 0.773 & 0.930 & 0.970 \\
DESC (Img. + Seg.) ~\cite{lopezrodriguez2020desc}                 & 0.174 & 1.480 & 5.818 & 0.250 & 0.771 & 0.918 & 0.966 \\
DESC (Img. + Seg. + Edge.) ~\cite{lopezrodriguez2020desc}   & \textbf{0.149} & 0.967 & 5.236 & 0.223 & \textbf{0.810} & 0.940 & 0.976 \\
DA-Monodepth (ours)           & 0.152 & \textbf{0.951} & \textbf{4.336} & \textbf{0.209} & 0.807 & \textbf{0.947} & \textbf{0.982} \\
\bottomrule[0.5mm]
\end{tabular}
\end{center}
\end{table*}


We perform our experiments on NVIDIA GTX Titan X GPU. In the training time, we provide source images, corresponding depth and semantic maps for Image encoder, Depth Decoder, and Semantic Decoder, but only provide source and target images for Domain Classifier. In the testing time, we provide target images as an input only, and depth maps as ground truth for evaluation. The evaluation are conducted on real world scenes as well as synthetic scenes, e.g., Virtual KITTI to KITTI or Cityscapes to KITTI, all of which contain outdoor scene structures but with different texture and lighting.

\subsection{\textbf{Implementation Details}} \label{Implementation Details}

Our network is built on the classic U-Net ~\cite{ronneberger2015unet} framework with the encoder-decoder structure and skip connections, leveraging the local and global features, allowing multi stage features concatenation to refine the final output. Similar to ~\cite{godard2019digging}, we choose ResNet-18 ~\cite{he2015deep} structure in our encoder, for its compactness and simplicity; we reverse the CNN channel order of ResNet-18 for our depth and semantic decoder accompany with bilinear upsampling in each decoder block. In terms of weight sharing, we only constrain the first decoder block of these two decoders to share weight and backpropagate the gradient together, since these two tasks are drifting apart as the decoder goes deeper. 

For the depth decoder, we utilize inverse depth(disparity) map as ground truth. In addition, we follow the same procedure as ~\cite{godard2019digging} to resize the multi-scale disparity maps to input images size, as we found that using low resolution disparity map to calculate loss and update the network will result in unnatural artifacts, such as holes and copied texture. These unnatural artifacts usually appears on low-texture area. After upsampling the low resolution disparity maps from multi-scale, we observe evident prediction improvement on low texture region. 

For the semantic decoder, we employ the same upsample strategy to calculate multi-scale cross entropy loss. Notice that we need to apply argmax operation channel-wise, to obtain the highest probability class for that pixel position. The semantic  ground truth label are store as integer for corresponding semantic class. In respect to shared block between depth decoder and semantic decoder, We average the summed gradient of two branches first, then preform the weight updating. 

As regard to domain classifier, we build this latent feature discriminator in a PatchGAN fashion ~\cite{isola2018imagetoimage}, which outputs a scalar patch for a given latent feature representation, and each value in that  patch holds account for a region in feature representation. This enables domain classifier to examine the input latent representation locally, hence lead to better discriminating ability compared to normal discriminator. With help of Gradient Reversal Layer, patch-based domain classifier can manipulate and align the latent feature representation efficiently and effectively.


\subsection{\textbf{Datasets}}
KITTI raw dataset~\cite{kittiraw} contains more than 40,000 image stereo pairs and sparse velodyne pointcloud. We choose Eigen split ~\cite{eigen2014depth} for our adaptation training, since it provides us a split with selected 22,600 training frames and 697 testing frames. The original resolution of KITTI image is $1242 \times 375$, and we resize the images to $640 \times 192$ before feeding the images into our network. 

Virtual KITTI dataset ~\cite{Gaidon:Virtual:CVPR2016} is a synthetic outdoor dataset with 21,260 photo-realistic images and corresponding depth and semantic segmentation maps. The maximum depth is 655.3 meters, thus we cap it to 80 meters to conform to KITTI settings.

Cityscapes~\cite{cordts2016cityscapes} is collected from real world with 5000 different urban images which include refined depth maps and semantic maps. The official split contains 2975 training image, 1525 testing images and 500 validation images. We follow the same procedure in ~\cite{lopezrodriguez2020desc} to center-crop and downsize the original image from $2048 \times 1024$ to $640 \times 192$.

\subsection{\textbf{Quantitative Results}}
Table~\ref{table1} showcases the adaptation results from \textbf{Virtual KITTI} to \textbf{KITTI}. We compare our DA-Monodepth model to the existing domain adaptation models on monocular depth estimation task. We evaluate all these methods on standard benchmarks using KITTI Eigen split~\cite{eigen2014depth} with median scaling. Our model demonstrates comparable results to DESC~\cite{lopezrodriguez2020desc}, the current state-of-the-art model, on all evaluation metrics. The full DESC model consists of an image translation module, instance segmentation module, and edge prediction module, in addition to the basic image encoder and depth decoder. In terms of training resources utilization, our model is similar to DESC model with image translation and instance segmentation module. In the comparison between DESC (Img Trans. + Seg.) and our DA-Monodepth model, we can clearly find the improvement for all benchmarks on either 80-meter or 50-meter caps. The same trends are also demonstrated in Table~\ref{table2}, where our DA-Monodepth model 
is comparable to the latest full DESC model and consistently outperforms the other models with the same data input settings.

\begin{table*}[t]
\begin{center}
\caption{Ablation study for \textbf{Virtual KITTI} to \textbf{KITTI} on the Eigen test split with median scaling. Baseline model is trained on source image and depth pair in a supervised fashion. For the second model, corresponding source segmentation map are introduced to model for shape representations, thereby improve the adaptation ability implicitly. For the last model, target image are added to aforementioned source data to complete the adaptation model.}
\label{table3}
\begin{tabular}{llllllll}
\toprule[0.5mm]
             & \multicolumn{4}{c}{Lower  $\downarrow$} & \multicolumn{3}{c}{Higher  $\uparrow$} \\
             \cmidrule(r){2-5}                        \cmidrule(l){6-8} 
             
Method      & Abs Rel &  Sq Rel &  RMSE &  RMSE log & $\delta < 1.25$  & $\delta < 1.25^{2}$ & $\delta < 1.25^{3}$      \\ 
\midrule[0.5mm]
\textbf{cap = 80m} \\
Source Img.+ Dep. (Baseline) & 0.206 & 2.186 & 7.366 & 0.270 & 0.762 & 0.913 & 0.959 \\
Source Img.+ Dep.+ Seg. & 0.190 & 2.186 & 6.794 & 0.254 & 0.769 & 0.920 & 0.968 \\
Source Img.+ Dep.+ Seg. + Target Img. & \textbf{0.156} & \textbf{1.272} & \textbf{5.118} & \textbf{0.221} & \textbf{0.783} & \textbf{0.927} & \textbf{0.972} \\

\midrule[0.5mm]
\textbf{cap = 50m} \\
Source Img.+ Dep. (Baseline) & 0.191 & 1.858 & 5.486 & 0.253 & 0.777 & 0.921 & 0.964 \\
Source Img.+ Dep.+ Seg. & 0.178 & 1.501 & 5.118 & 0.239 & 0.783 & 0.927 & 0.972 \\
Source Img.+ Dep.+ Seg. + Target Img.  & \textbf{0.149} & \textbf{0.958} & \textbf{4.321} & \textbf{0.210} & \textbf{0.811} & \textbf{0.947} & \textbf{0.982} \\

\bottomrule[0.5mm]
\end{tabular}
\end{center}
\end{table*}

\begin{table*}[t]
\begin{center}
\caption{Ablation study for for \textbf{Cityscapes} to \textbf{KITTI} on the KITTI Eigen test split with median scaling.}
\label{table4}
\begin{tabular}{llllllll}
\toprule[0.5mm]
             & \multicolumn{4}{c}{Lower  $\downarrow$} & \multicolumn{3}{c}{Higher  $\uparrow$} \\
             \cmidrule(r){2-5}                        \cmidrule(l){6-8} 
             
Method      & Abs Rel &  Sq Rel &  RMSE &  RMSE log & $\delta < 1.25$  & $\delta < 1.25^{2}$ & $\delta < 1.25^{3}$      \\ 
\midrule[0.5mm]
\textbf{cap = 80m} \\
Source Img.+ Dep. (Baseline)                 & 0.216 & 3.276 & 5.891 & 0.277 & 0.772 & 0.926 & 0.971 \\
Source Img.+ Dep.+ Seg.    & 0.163 & 1.064 & 4.711 & 0.218 & 0.776 & 0.946 & 0.983 \\
Source Img.+ Dep.+ Seg. + Target Img.          & \textbf{0.152} & \textbf{0.951} & \textbf{4.336} & \textbf{0.209} & \textbf{0.807} & \textbf{0.947} & \textbf{0.983} \\
\bottomrule[0.5mm]
\end{tabular}
\end{center}
\end{table*}
\subsection{\textbf{Ablation Study}}

In Table~\ref{table3} and Table~\ref{table4}, we compare the performances of different module combinations which described in Section~\ref{Implementation Details}. As for the baseline model, it only contains a Image encoder and depth decoder, and only has access to source RGB image and corresponding depth. For the second model (Source Img. + Dep.+ Seg), a semantic segmentation decoder with shared weight is added to the baseline model, and we feed source semantic segmentation map additionally. The last model (Source Img. + Dep.+ Seg.+ Target Img.) is the complete model with an extra domain classifier than the second one, we feed all the aforementioned source data as well as target RGB images. 

Comparing the baseline model and second model on standard depth estimation metrics, the second one consistently outperforms over baseline model. Since the semantic information helps the model to learn more shape-based features instead of texture-base features, introducing semantic information benefits the domain adaptation process.
For the last model, the metrics obviously shows the advantage of using domain classifier to narrow the domain gap with respect to baseline model, and improves the performance on the base of second model. Hence, semantic segmentation decoder and domain classifier have demonstrated the ability to bridge the domain discrepancy during training process. 

\subsection{\textbf{Qualitative Results}}
\begin{figure*}[ht]
      \center
      \includegraphics[width=0.87\linewidth]{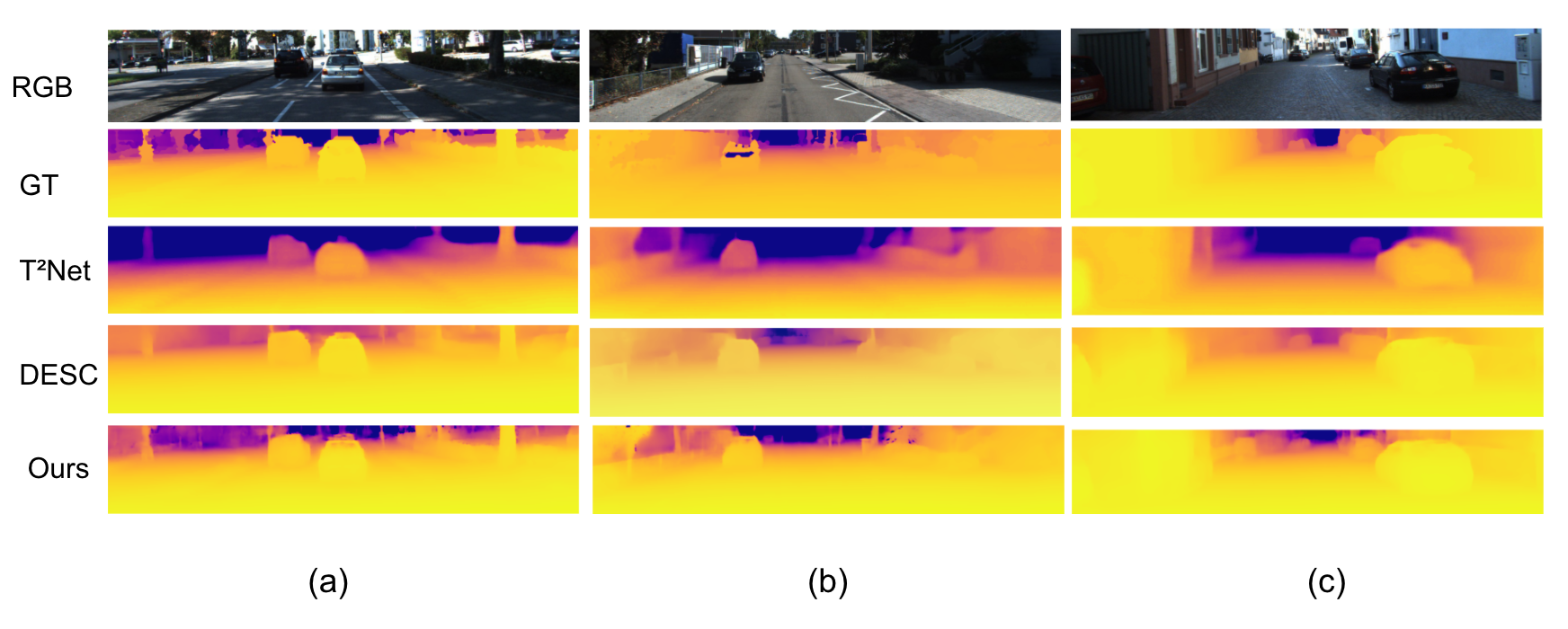}
      \caption{Qualitative adaptation results from \textbf{Virtual KITTI} to \textbf{KITTI}. GT represents ground truth that manually interpolated from sparse LiDAR pointcloud. All results are cropped to ground truth shape for better visualization.}
      \label{figure4}
\end{figure*}

\begin{figure*}[ht] 
      \center
      \includegraphics[width=0.87\linewidth]{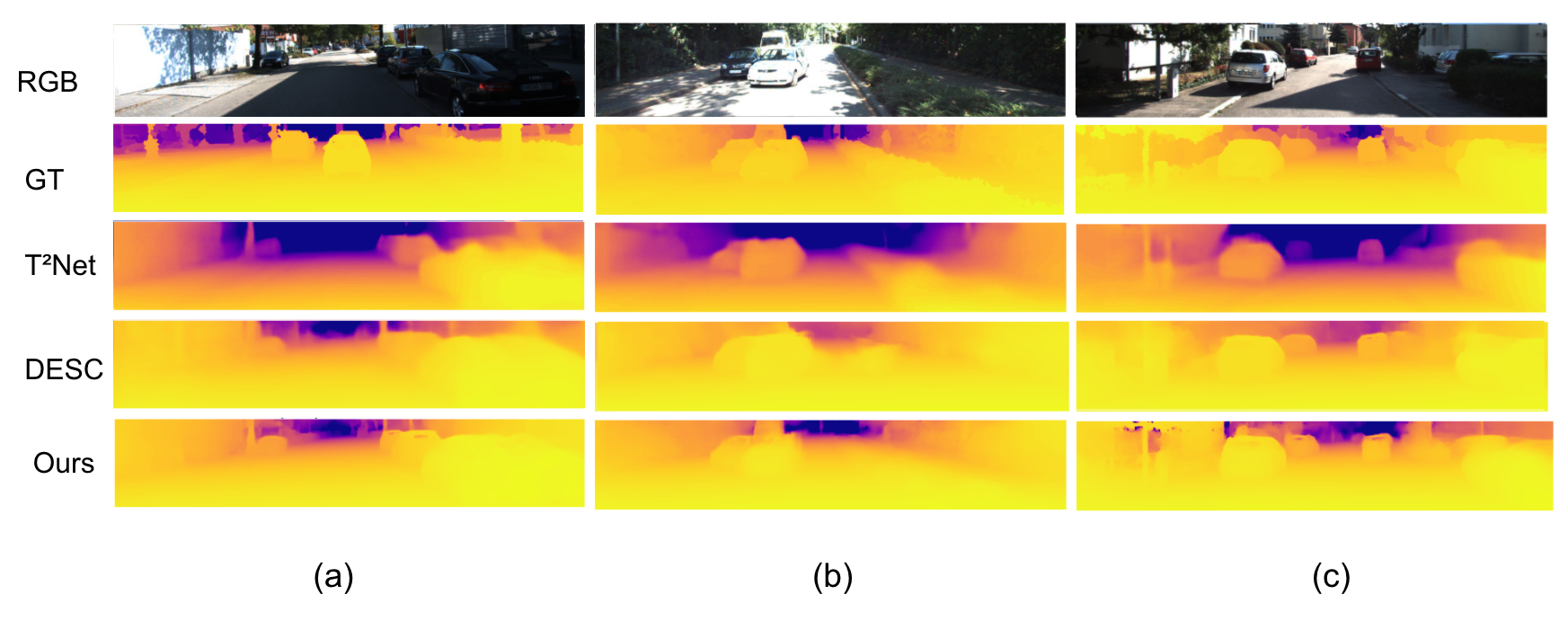}
      \caption{Qualitative adaptation results for models trained in \textbf{Cityscapes} to \textbf{KITTI}. }
      \label{figure5}
\end{figure*}

Figure~\ref{figure4} demonstrated our model's adaptation ability against the previous state-of-the-art models qualitatively. All of the models were trained in \textbf{Virtual KITTI} to \textbf{KITTI} settings. We manually interpolated ground-truth depth from sparse depth map using a sliding window and a bilinear filter. Generally, T\textsuperscript{2}Net approximately recovers the near-field depth and object shapes with blurry edges. DESC visually improves the fine-grained details that are largely missed in T\textsuperscript{2}Net due to the added instance mask information. Our model outperforms above models by recovering more details in those structures appear far away or in shadowed areas. For example, in Figure \ref{figure4} column (a), T\textsuperscript{2}Net ignores details from a distance resulting in more purple regions in the background; DESC performs better on the poles and bushes but estimates the background depth shorter than the ground truth. In Figure \ref{figure4} column (c), our model picks up even the small cars in the scene as well as accurately predicting the background depth. Figure \ref{figure4} also shows consistent results with all models adapted from \textbf{Cityscapes} to \textbf{KITTI}. In Figure \ref{figure5} column (b), only our model captures the shape of a van behind the cars; In Figure \ref{figure5} column (c), our model predicts the finest depth details including foreground trees and background buildings.

\subsection{\textbf{Training Complexity}}
\begin{table}[ht]
\begin{center}
\caption{Adaptation complexity for \textbf{Virtual Kitti} to \textbf{KITTI} using the KITTI Eigen split. All models are trained on 22,600 images for 20 epochs.}
\begin{tabular}{ll}
\toprule[0.5mm]
Method & Trainable Parameters \\
\midrule[0.3mm]
T\textsuperscript{2}Net & 54.6 M  \\
DESC                    & 55M  \\
DA-Monodepth            & 21.5M  \\
\bottomrule[0.5mm]
\end{tabular}
\end{center}
\label{table5}
\end{table}

Table \ref{table5} shows the trainable parameters of all these methods. Since DESC builds on T\textsuperscript{2}Net, both models share the same backbone structure consisting of an image translation module and a depth prediction module, albeit with different auxiliary modules; thereby, the two models contain similar numbers of trainable parameters. Our model, DA-Monodepth, is based on a simplest U-net variant with minimal compact auxiliary modules, namely semantic decoder and domain classifier, drastically reducing the number of parameters to train. Meanwhile, our model outputs comparable results against complex state-of-the-art models such as DESC and generalizes better on target domains compared to the recent domain adaptation models.

\section{\textbf{Conclusions}}

We propose a compact, semi-supervised domain adaptation model \textbf{Domain Adaptive Monodepth}, aiming to tackle the monocular depth estimation problem. Our model leverages semantic information and adversarial training to jointly improve the cross-domain generalization ability and the depth prediction accuracy. To bridge the domain gap, we introduce a domain adaptation module that includes a Gradient Reversal Layer (GRL) and a domain classifier together to minimize the domain discrepancy and align latent feature representations. To recover the fine-grained details and scene structures in depth map prediction, we also incorporate a semantic segmentation module. We conduct experiments on \textbf{Virtual KITTI} to \textbf{KITTI} and \textbf{Cityscapes} to \textbf{KITTI} settings, and demonstrate that our model performs comparably against the state-of-the-art models on the KITTI depth prediction benchmark. Qualitatively, our model shows advantages in estimating depths for small-sized, far-away objects as well as clear object boundaries. Ablation studies show that the multi-task setting benefits latent representations and that adversarial learning is effective in aligning latent representations. Lastly but importantly, our model reaches the state-of-the-art performance with a relatively compact design that helps speed up adaptation training.

{\small
\bibliographystyle{ieee_fullname}

}

\end{document}